\documentclass[3p,times,procedia]{elsarticle}
\usepackage{ecrc}
\usepackage[bookmarks=false]{hyperref}
    \hypersetup{colorlinks,
      linkcolor=blue,
      citecolor=blue,
      urlcolor=blue}
\usepackage{xcolor}
\usepackage{colortbl}
\usepackage{makecell}
\usepackage{ulem}

\volume{00}

\firstpage{1}

\journalname{Procedia Computer Science}

\runauth{Author name}

\jid{procs}

\usepackage{amssymb}
\usepackage{amsmath,amssymb,amsfonts}
\usepackage{booktabs}
\usepackage{multirow}
\usepackage{hyperref}

\usepackage[figuresright]{rotating}

\begin{document}
\begin{frontmatter}



\dochead{30th International Conference on Knowledge-Based and Intelligent Information \& Engineering Systems (KES 2026)}%

\title{Automated Compliance Mapping in Cloud Security with Domain-Adapted Sentence Transformers}



\author[2,1]{John Bianchi\fnref{*}}
\author[1,2]{Luca Petrillo\fnref{*}}
\author[3]{Fabio Martinelli}
\author[1,2]{Marinella Petrocchi}

\fntext[*]{These authors contributed equally to this work.}

\address[1]{Institute for Informatics and Telematics (IIT-CNR), Pisa, Italy}
\address[2]{IMT School for Advanced Studies Lucca, Lucca, Italy}
\address[3]{Institute for High Performance Computing and Networking (ICAR-CNR), Rende (CS), Italy}

\begin{abstract}
Mapping cloud security controls to technical metrics is currently a manual process. This paper proposes domain adaptation of Sentence Transformer models to automate it. We build a training corpus of 3,499 semantic pairs from five European security standards and a set of technical metrics, then expand it via back-translation and LLM-based paraphrasing to up to 13,996 samples across four scenarios. We fine-tune five architectures and evaluate their performance on two independent tasks: control-to-metric and cross-standard controls association. All fine-tuned models outperform their zero-shot baselines. On the control-to-metric task, the best model gains up to 23 nDCG@10 points, while on the cross-standard control task, \textit{multi-qa-mpnet-dot-v1} under back-translation reaches 0.870 nDCG@10. The results show that in-domain training data is a primary driver of performance for the considered case studies.\end{abstract}

\end{frontmatter}



\section{Introduction}

Cloud Service Providers (CSPs) in Europe operate within a shifting regulatory environment. The European Cybersecurity Certification Scheme for Cloud Services (EUCS)~\cite{ENISA2020} aims to replace fragmented national standards with a unified framework. While this unification strengthens security, it imposes a heavy compliance burden on providers.
To obtain certification, CSPs must demonstrate that their technical operations satisfy high-level legal mandates. This process relies on technical evidence, which we call \textit{Metrics}. Metrics support the evaluation of abstract normative texts, which we call \textit{Controls}.

Currently, experts carry out the mapping between controls and metrics manually. This process is slow, costly and error-prone. The continuous evolution of standards (e.g., the transition from BSI C5 to EUCS \cite{BSIc5}) means organizations have to redo these mappings repeatedly, creating a scalability bottleneck.
Existing approaches to compliance automation leave a clear gap. The efforts mainly focus on structured formats, while cross-standard mapping and compliance reasoning still rely heavily on manual modeling and rule-based approaches, often requiring handcrafted ontologies or metamodels that are brittle to control rephrasing \cite{castellanos2022compliance}. At the same time, semantic similarity models based on Sentence Transformers perform well on general text, yet they are not specifically trained on cloud-security language, nor systematically evaluated for both control-to-metric and control-to-control matching.

In our previous work~\cite{emerald1, emerald2}, we explored the use of pre-trained Sentence Transformers (SBERT) to match controls to metrics. That study relied on generic models with no prior knowledge of cloud security terminology. Although the results showed promise ($nDCG_{10} = 0.640$), performance remained limited, likely due to the absence of domain-specific knowledge.
This paper addresses this gap through domain adaptation via fine-tuning, using a data-centric approach for cloud-compliance matching. In particular, the main contributions of this work are:

\begin{itemize}
    \item We introduce a domain-specific training corpus for cloud compliance, built by integrating heterogeneous standards (BSI C5, ENS, SecNumCloud, EUCS) through a common reference framework (Cisco CCF) and associating them with metrics developed within the EMERALD EU project~\cite{emeraldProject}.
    
    \item We fine-tune Sentence Transformer models on this corpus to support both control-to-metric and control-to-control matching tasks in cloud-security compliance.
    
    \item We investigate the impact of data augmentation strategies, including back-translation and LLM-based paraphrasing, across four training scenarios.
    
    \item We show that domain-adapted models perform better than generic baselines, demonstrating the importance of domain-specific knowledge for automated compliance mapping.
\end{itemize}

\vspace{-0.75cm}


\section{Related Work}


The use of NLP for regulatory compliance has been explored in several settings. Chalkidis et al.~\cite{chalkidis2021regulatory} treated cross-legislation alignment as a document retrieval problem over EU and UK statutes, showing that in-domain fine-tuning improves retrieval over lexical baselines when surface-level text similarity is weak. The work in~\cite{ciaramella2025leveraging} leveraged pre-trained transformer models to determine whether a privacy policy complies with the single duty specified in Article 13(2)(b) of the GDPR, i.e., whether it informs data subjects of their right to rectification or erasure of personal data. Amaral et al.~\cite{cejas2023nlp} applied NLP to check data processing agreements against GDPR requirements automatically, targeting the semantic gap between legal mandates and operational artifacts. Carello et al.~\cite{carello2024study} mapped cybersecurity controls to vulnerability catalogs using semi-automated NLP techniques, confirming that vocabulary mismatch between normative and technical text remains an open problem in the security domain. 
In~\cite{agarwal2021ai}, Agarwal et al. map regulatory controls to NIST 800-53 using fine-tuned Transformers (best: hierarchical classification, Recall@K on 1,580→826 mappings). In~\cite{ahmed2024prompting}, Ahmed et al. translate Center for Internet Security (CIS) Critical Security Controls into measures/metrics via few-shot LLM prompting, evaluated by LLM-as-judge correlation. Unlike both, we target EU schemes (EUCS, BSI C5, ENS, SecNumCloud) and jointly address control-to-control and control-to-metric retrieval with domain-adapted sentence transformers.

In our previous work~\cite{emerald1, emerald2}, we evaluated pre-trained sentence transformers for the association of cloud security controls with metrics in a zero-shot setting. We found that generic models performed modestly without domain knowledge. Building on this research, this study creates a labeled training corpus from various European standards and uses back-translation and LLM-based paraphrasing as augmentation strategies. It also examines the impact of fine-tuning across four training configurations.

\section{Data Sources}
\label{data_sources}
We compile our knowledge base from six sources. These sources cover three dimensions: structural association via a central hub, technical specificity via metric catalogues and linguistic variety via national standards. Two additional datasets, the MEDINA metric catalogue and the EUCS-Legacy control snapshot,  are reserved exclusively for evaluation purposes and are described in Section~\ref{sec:eval_sets}.

\paragraph{Cisco Cloud Controls Framework}
\label{ccf}
The CISCO Cloud Controls Framework (CCF) v2.0~\cite{ciscoCCF} is a publicly available framework that aggregates international and national security compliance and certification requirements, including ISO 27001\footnote{ISO/IEC 27001 – Information security management systems standard: \url{https://www.iso.org/standard/27001}} and SOC 2\footnote{Service Organization Control 2 – Trust services criteria for security, availability, and confidentiality: \url{https://www.aicpa-cima.com/topic/audit-assurance/audit-and-assurance-greater-than-soc-2}}, into a single unified resource. It provides a structured approach to managing cloud security controls and monitoring their effectiveness in cloud infrastructure deployments. The CCF encompasses multiple control domains, including identity and access management, data protection, incident response, and vulnerability management. Each domain contains specific controls that organizations can implement based on their risk profile and regulatory requirements. 

\paragraph{Cloud Computing Compliance Criteria Catalogue}
The Cloud Computing Compliance Criteria Catalogue (BSI C5)~\cite{BSIc5}, developed by the German Federal Office for Information Security, is a set of security and compliance criteria specifically designed for evaluating cloud service providers. This catalogue aims to establish minimum security standards and requirements that cloud providers must adhere to, ensuring the protection of sensitive data and critical infrastructure in cloud environments. It specifically encompasses thirteen primary control areas, including information security management, identity and access management, cryptography and key management, where each domain contains detailed criteria and requirements that cloud service providers must satisfy to achieve certification. In the context of this work, it serves as a structural bridge in our knowledge base as it is the only source with native mappings both to the CCF and to the EMERALD operational metrics, which connects normative requirements to technical evidence.

\paragraph{Esquema Nacional de Seguridad}
The Esquema Nacional de Seguridad (ENS)~\cite{spanishENS} is a Spanish regulatory framework that establishes principles and requirements for the security of information handled by the Spanish public sector entities, aiming to ensure the protection of information systems and data. It is built on the CIA triad (confidentiality, integrity, and availability) and mandates a risk-based approach, requiring organizations to conduct security assessments to implement proportionate security controls.

\paragraph{SecNumCloud}
SecNumCloud (v3.2) \cite{secnumcloud} is a French national cloud security qualification framework established by the French National Cybersecurity Agency (ANSSI)\footnote{https://cyber.gouv.fr}. It defines stringent security, compliance, and data sovereignty requirements for cloud service providers handling sensitive data, particularly for public sector and critical use cases. The framework applies to providers offering services under European jurisdiction, ensuring protection against extra-territorial laws and strong guarantees of data sovereignty, often involving infrastructure located within France or the EU.

\paragraph{EUCS Candidate Scheme}
The European Cybersecurity Certification Scheme for Cloud Services (EUCS) (draft version v2020~\cite{ENISA2020}) is a framework proposed to standardize cybersecurity certifications across the European Union in order to harmonize national certification schemes, security, and governance standards for cloud service providers. 

\paragraph{Repository of Controls and Metrics}
The Horizon Europe EMERALD Project~\cite{emeraldProject} is developing an evidence management platform for continuous certification-as-a-service in the cloud. This includes a repository of controls and metrics for assessing compliance with schemes such as those listed in previous subsections. The security metrics cover various areas, including transport encryption, TLS-related metrics, password metrics and antimalware scan frequency. 

\section{Problem Definition}
We formalize the association between controls from different standards, as well as between controls and metrics, as an open-domain information retrieval problem. Keyword search is insufficient here because the vocabulary and abstraction level of controls and metrics, as well as controls from different standards, differ. A technical metric such as \textit{``OPS-05.3H: Antimalware enabled''} and a normative control such as \textit{``The CSP shall deploy malware detection mechanisms on all compute nodes''} express the same requirement in very different terms.
Let $\mathcal{Q}$ be a set of queries and $\mathcal{D}$ be a set of documents.  Our system takes a control as input, thus a query $q \in \mathcal{Q}$ is always a control. Conversely, a list of documents $d \in \mathcal{D}$ represents the target artifacts to be retrieved, which can be either controls from a target scheme (e.g., EUCS) or technical metrics.

The goal is to learn a similarity function $f(q, d) \rightarrow \mathbb{R}$ (where $\mathbb{R}$ is the set of real numbers representing the similarity score) that is used to score a set of candidate documents. Given a query $q$, the system applies $f$ to produce a ranked list of documents $d \in \mathcal{D}$ ordered by decreasing similarity scores, such that the documents semantically satisfying $q$ appear at the top of the ranking.

We parameterize $f$ using a bi-encoder based on Sentence 
Transformers, which maps $q$ and $d$ independently into a shared 
vector space:

\begin{equation}
    f(q, d) = \cos(\mathbf{u}, \mathbf{v}) = 
    \frac{\mathbf{u} \cdot \mathbf{v}}{\|\mathbf{u}\| \|\mathbf{v}\|}
\end{equation}

where $\mathbf{u}$ and $\mathbf{v}$ represent the embeddings extracted from $q$ and $d$, respectively.

The model evaluates a pair of texts $(q, d)$ in input and outputs a scalar similarity score, without having a built-in notion of whether the text represents a metric or a control. While the system's final output is a ranked list, the underlying function operates on pairs, learning that some pairs are semantically related and others are not. A single formulation allows control-to-control and control-to-metric pairs to contribute to the same learning objective, resulting in a larger, more uniform training corpus. It also eliminates the need to maintain separate models for each task, reducing engineering complexity and making the system easier to extend to new standards.
Thus, this formulation covers two tasks. In \textbf{Control-to-Control Association}, the query is a control from one framework and the goal is to retrieve semantically equivalent controls from another (e.g., a control in CCF to a control in EUCS). In 
%
\textbf{Control-to-Metric Association}, the query is a control, and the goal is to retrieve the metrics that declines that control in a technical way.

\section{Methodology}

Pre-trained generic models can underperform on specialized regulatory texts~\cite{emerald1, emerald2}. To address this issue, we have adopted a three-stage domain adaptation pipeline. First, we construct a corpus of semantic pairs from our knowledge base. Second, we expand this corpus through data augmentation. We then fine-tune the encoder using a contrastive learning objective~\cite{gao2021simcse}.
Given a query $q$, let $d^+$ denote a relevant (positive) document and $d^-$ an irrelevant (negative) one. The objective encourages representations of $(q, d^+)$ to be close in the embedding space, while pushing $(q, d^-)$ farther apart.

\subsection{Training Corpus Construction}
We construct a labeled dataset of anchor–target pairs, where each pair links a control from one framework to a semantically equivalent control (or metric) from another. The dataset is assembled from publicly documented frameworks, as described in Section 3, using two different strategies.

\subsubsection{Semantic Pair Extraction}
\paragraph{Strategy I: Control-to-Control Association via Cisco CCF}
As described in Section \ref{ccf}, the Cisco CCF serves as a central hub for various international and national security compliance requirements and has also been designed to provide cross-framework mapping. In particular, ENS, BSI C5, and SecNumCloud have mappings to CCF, so we can use it to derive mappings between the three schemes as well. 

Since both ENS and SecNumCloud are distributed as unstructured PDFs in their source languages (Spanish and French), we use \textit{pdfplumber}\footnote{\url{https://github.com/jsvine/pdfplumber}} to extract the text from these documents, and we subsequently translate them into English using the DeepL API\footnote{\url{https://www.deepl.com/pro-api}}. 
 
As the Cisco CCF dataset contains columns that indicate which ENS controls correspond to each CCF control, we extract the referenced ENS codes and match them against a lookup table that was created using the ENS dataset. This results in 318 pairs linking ENS controls to CCF controls. We apply the same procedure to SecNumCloud, obtaining an additional 563 pairs. We then map the CCF dataset directly to BSI C5 pairs, resulting in an additional 555 pairs. Similarly, for the mapping between CCF and EUCS-2020 (the 2020 draft candidate scheme~\cite{ENISA2020})\footnote{We would like to highlight that EUCS-2020 differs from the EUCS-Legacy version used in evaluation, see Section~\ref{sec:eval_sets}.}, we extract the mappings directly from the Cisco CCF dataset, producing 1,270 pairs. Finally, we create a further 564 direct EUCS-2020-to-BSI-C5 mappings.

\paragraph{Strategy II: Control-to-Metric Association via Semantic Bridge}
The EMERALD partners are working on constructing a catalogue of metrics to create pairs connecting the metrics to controls. As the EMERALD catalogue contains references to BSI C5 controls, we can match the metrics directly against the BSI C5 lookup table to yield 27 pairs.
The catalogue does not provide a direct mapping to the EUCS, but we have the mapping between the EUCS and C5 controls, so we can also determine the associations between the catalogue's metrics and EUCS (202 new semantic pairs).

The semantic pairs generated through Strategies I and II form a dataset of 3,499 pairs. This collection forms the base training corpus (\textbf{Scenario I}) prior to any data augmentation.

\subsubsection{Data Augmentation}
\label{sec:data_aug}
We apply two augmentation strategies. The back-translation strategy adds 3,499 samples, while the LLM-based paraphrasing, using two models, adds 6,998 samples. This results in three additional scenarios:

\begin{itemize}
    \item \textbf{Scenario II} --- Original + Back-Translation: 6,998 samples
    \item \textbf{Scenario III} --- Original + Paraphrase: 10,497 samples
    \item \textbf{Scenario IV} --- Original + Back-Translation + Paraphrase: 13,996
    samples
\end{itemize}

\paragraph{Back-Translation}
We translate each text into a randomly selected intermediate language using the Google Translate API, then translate it back to English. This introduces lexical and syntactic variation while preserving meaning. 

\paragraph{LLM-Based Paraphrasing}
We use two instruction-tuned models to paraphrase each text:

\begin{itemize}
    \item \textit{microsoft/Phi-4-mini-instruct\footnote{\url{https://huggingface.co/microsoft/Phi-4-mini-instruct}}}
    \item \textit{meta-llama/Llama-3.1-8B-Instruct\footnote{\url{https://huggingface.co/meta-llama/Llama-3.1-8B-Instruct}}}
\end{itemize}

Both models run in 4-bit quantization (NF4) via \texttt{BitsAndBytesConfig}. Each model receives a prompt instructing it to rephrase the input while preserving: (i) the technical meaning, (ii) domain-specific terminology (e.g., \textit{least-privilege}, \textit{two-factor authentication}), and (iii) the logical structure of the original text. 
The outputs of both models are pooled into a single paraphrase set, contributing 6,998 samples in Scenarios~III and~IV.

The proportion of pairs from each source-target mapping remains consistent throughout the augmentation process, ensuring that no mapping type is over-represented in the final augmented corpus.


\begin{table}[h]
\caption{Composition of the training and test datasets. The training block lists the number of pairs per mapping topology. The augmentation block shows the contribution of each strategy and the total sample count per experimental scenario. Both test sets are held out across all four scenarios.}
\label{tab:dataset_stats}
\centering
\resizebox{\columnwidth}{!}{\begin{tabular}{llllrrr}
\toprule
\textbf{Phase} & \textbf{Topology} & \textbf{Anchor Source} ($A$) &
\textbf{Positive Target} ($P$) & \textbf{Unique} $|A|$ &
\textbf{Unique} $|P|$ & \textbf{Pairs} $(A,P)$ \\
\colrule
\multirow{8}{*}{\textbf{Training}}
 & Hub Association             & Cisco CCF       & Spanish ENS              & 713 & 169 & 318   \\
 & Hub Association             & Cisco CCF       & SecNumCloud              & 713 & 261 & 563   \\
 & Hub Association             & Cisco CCF       & BSI C5:2020              & 713 & 223 & 555   \\
 & Hub Association             & Cisco CCF       & EUCS-2020                & 713 & 522 & 1,270 \\
 & Semantic Bridge           & BSI C5:2020               & Emerald-Metrics & 27  & 57  & 27    \\
 & Semantic Bridge           & EUCS-2020 (via BSI)       & Emerald-Metrics & 110 & 57  & 202   \\
 & Cross-Standard (Filtered) & EUCS-2020       & BSI C5:2020              & 489 & 107 & 564   \\
\cmidrule{2-7}
 & \multicolumn{3}{r}{\textit{Scenario I --- Original (after augmentation)}}
   & \multicolumn{3}{r}{\textbf{3,499}} \\
\midrule
\multirow{6}{*}{\textbf{Augmentation}}
 & Back-Translation  & Google Translate      & (all pairs) & \multicolumn{3}{r}{+3,499 \hspace{1.3em}} \\
 & Paraphrase        & Phi-4-mini-instruct   & (all pairs) & \multicolumn{3}{r}{+3,499 \hspace{1.3em}} \\
 & Paraphrase        & Llama-3.1-8B-Instruct & (all pairs) & \multicolumn{3}{r}{+3,499 \hspace{1.3em}} \\
\cmidrule{2-7}
 & \multicolumn{3}{r}{\textit{Scenario II --- Original + Back-Translation}}
   & \multicolumn{3}{r}{\textbf{6,998}} \\
 & \multicolumn{3}{r}{\textit{Scenario III --- Original + Paraphrase}}
   & \multicolumn{3}{r}{\textbf{10,497}} \\
 & \multicolumn{3}{r}{\textit{Scenario IV --- Original + Back-Translation + Paraphrase}}
   & \multicolumn{3}{r}{\textbf{13,996}} \\
\midrule
\multirow{2}{*}{\textbf{Test}}
 & \textbf{Test Set A} (Control-to-Metric) & EUCS-Legacy & Medina-Metrics & 70  & 166 & 179 \\
 & \textbf{Test Set B} (Cross-Standard)    & EUCS-Legacy    & BSI C5:2020 & 140 & 51  & 140 \\
\botrule

\end{tabular}}
\end{table}

\subsection{Dataset Statistics}

Table~\ref{tab:dataset_stats} shows the full corpus structure. The CCF hub produces the majority of training pairs by linking its controls to four target standards. The EMERALD bridge adds a smaller set of pairs that controls to metrics. Then we have the cross-standard set that maps EUCS-2020 to C5:2020. The seven mapping steps produce 3,499 pairs before augmentation. 
Each anchor can be associated with one or more targets (one-to-many relationship), reflecting that a single normative control may be linked to multiple equivalent controls or technical metrics. The unique counts for the Cross-Standard mapping (Row 7) are lower than the full EUCS-2020 and BSI C5:2020 pool sizes because EUCS-2020 entries overlapping with the test sets were removed by the decontamination filter. The 564 pairs are generated as follows: each EUCS-2020 control may reference multiple BSI C5 controls, and each such reference yields one training pair. Only references that survive the decontamination filter and match a known BSI C5 entry are retained. For example, a single EUCS control mapped to three BSI C5 controls contributes three distinct pairs to the corpus.

\subsection{Test Sets}
\label{sec:eval_sets}
We evaluate all models on two test sets that are completely independent of any training scenario.

\paragraph{Test Set A --- Control-to-Metric Association}
This set comes from the MEDINA project~\cite{medinaProject}, a Horizon 2020 European project predecessor of Emerald. The queries comprise 70 unique controls from a specific snapshot of the EUCS 2020 candidate scheme, referred to as EUCS-Legacy\footnote{EUCS-Legacy is a subversion of the 2020 candidate scheme adopted during the early stages of the MEDINA project, 
sharing the same identifier and file structure as the official 2020 release but featuring minor structural and phrasing differences from it.}. The target corpus consists of 166 unique metrics defined by the MEDINA partners, where each metric is a short identifier-prefixed technical description (e.g., \textit{``OPS-05.3H: Antimalware enabled''}). The ground truth, that is, the set of control-metric pairs, comprises 179 pairs produced by domain experts. 

\paragraph{Test Set B --- Cross-Standard Control-to-Control Association}
The queries are 140 EUCS-Legacy controls, while the target corpus is 51 BSI C5:2020 controls. The ground truth comprises 140 pairs derived by mapping EUCS-Legacy identifiers to the corresponding BSI C5 controls. 

\paragraph{Decontamination}\label{sec:decontamination}
We filter the training corpus to remove any EUCS-2020 controls whose text overlaps with an EUCS-Legacy entry in either of the test sets.  This filter is applied before augmentation, meaning that no back-translated or paraphrased variant of a filtered entry will be included in any training scenario.

\subsection{Model Selection}
We evaluate the same five Sentence Transformer models used in~\cite{emerald1, emerald2}. This allows for a direct comparison of zero-shot and fine-tuned performance on the same architectures, thus eliminating the confounding factor of model selection.  The five models differ in terms of their architecture, size and pre-training objective, and they encompass both general-purpose encoders and retrieval-oriented variants. This allows us to determine whether domain adaptation improves performance consistently across encoder families or if its impact is architecture-specific.
\begin{itemize}
    \item \textit{all-mpnet-base-v2}\footnote{\url{https://huggingface.co/sentence-transformers/all-mpnet-base-v2}}: A general-purpose MPNet encoder trained on diverse text pairs. It maps sentences to a 768-dimensional vector space and uses cosine similarity for scoring.

    \item \textit{multi-qa-mpnet-base-dot-v1}\footnote{\url{https://huggingface.co/sentence-transformers/multi-qa-mpnet-base-dot-v1}}: An MPNet variant trained for question-answering retrieval across multiple domains. It uses dot-product scoring instead of cosine similarity.

    \item \textit{all-distilroberta-v1}\footnote{\url{https://huggingface.co/sentence-transformers/all-distilroberta-v1}}: A distilled RoBERTa encoder with fewer parameters than the MPNet models. It has faster inference and a different base architecture.

    \item \textit{all-MiniLM-L12-v2}\footnote{\url{https://huggingface.co/sentence-transformers/all-MiniLM-L12-v2}}: A compact encoder with 12 layers and a reduced hidden size. It performs competitively on general semantic similarity benchmarks.

    \item \textit{multi-qa-distilbert-cos-v1}\footnote{\url{https://huggingface.co/sentence-transformers/multi-qa-distilbert-cos-v1}}: A DistilBERT encoder trained for retrieval on multi-domain question-answering data. It uses cosine similarity and a different base architecture from the MPNet
    models.
\end{itemize}

\subsection{Baseline}

We evaluate each model on both test sets prior to any fine-tuning, using the models' original off-the-shelf weights. These performances serve as the baseline that domain-adapted models must improve upon. These also replicate the zero-shot setting of~\cite{emerald1, emerald2}, enabling us to evaluate the benefits of domain adaptation in relation to the previous study.

\subsection{Training Strategy}
We fine-tune each of the five models across the four training scenarios, yielding 20 fine-tuned models in total. The training objective is Multiple Negatives Ranking Loss (MNRL)~\cite{gao2021simcse}. For each positive pair in a batch, all other positive pairs in the same batch act as implicit negatives. The model learns to score the correct pair higher than all others without requiring manually labeled negative examples. All runs use a batch size of 64, 3 training epochs, a learning rate of $2 \times 10^{-5}$, and a linear warmup over the first 10\% of training steps.

\subsection{Evaluation Protocol}
At inference time, the model operates in two directions. For Test Set~A, controls are the queries and metrics are the candidate pool. For Test Set~B, controls from one framework are the queries and controls from another framework are the candidate pool. In both cases, the model computes cosine similarity against every candidate and returns a ranked list. The model has no notion of whether a candidate is a metric or a control. It scores all text pairs with the same function. This retrieval direction matches the setting in~\cite{emerald1, emerald2} and reflects the operational use case: an auditor starts from a normative control and retrieves the artifacts that satisfy it or similar controls. We evaluate the ranking quality using nDCG@10.

\paragraph{Normalized Discounted Cumulative Gain (nDCG@10)}
nDCG@10 measures ranking quality up to rank 10, giving higher importance to relevant items appearing at top positions through logarithmic discounting.
\begin{equation*}
\text{DCG}@k = \sum_{i=1}^k \frac{rel_i}{\log_2(i+1)},
\qquad
\text{nDCG}@k = \frac{\text{DCG}@k}{\text{IDCG}@k}
\end{equation*}

Here, $rel_i \in {0,1}$ denotes the relevance of the item at rank $i$, and IDCG@$k$ corresponds to the maximum possible DCG obtained from an ideal ranking. 

\section{Results}
We evaluate all 25 models (5 architectures $\times$ 4 training variants, plus 5 baselines) on both test sets.  The results are reported as nDCG@10 for all queries, as well as for queries with at least one relevant document among the top 10 results. These results are shown in Table~\ref{tab:results_fine_tuning}, where the first two rows are the zero-shot baselines for each architecture. Throughout the table, the best performance per architecture is highlighted in bold, while the best result across all models is in bold and underlined.

\newcommand{\headercell}[1]{\textbf{\scriptsize\makecell[c]{#1}}}
\begin{table}[ht]
\caption{Performance comparison (nDCG@10). \textbf{Bold} values indicate the best score per architecture, while \underline{\textbf{underlined bold}} indicates the best score across all models. A\,=\,Test Set A (Control-to-Metric Association); B\,=\,Test Set B (Cross-Standard Control Association); DA\,=\,Dataset.}
\label{tab:results_fine_tuning}
\centering
\scriptsize
\renewcommand{\arraystretch}{1.1} 
\setlength{\tabcolsep}{2pt} 

\resizebox{\textwidth}{!}{%
\begin{tabular}{lccccccccccc}
\toprule
 & & \multicolumn{2}{c}{\headercell{all-mpnet-\\base-v2}} & \multicolumn{2}{c}{\headercell{multi-qa-mpnet-\\base-dot-v1}} & \multicolumn{2}{c}{\headercell{all-distil\\roberta-v1}} & \multicolumn{2}{c}{\headercell{all-MiniLM-\\L12-v2}} & \multicolumn{2}{c}{\headercell{multi-qa-\\distilbert-cos-v1}} \\
\cmidrule(lr){3-4} \cmidrule(lr){5-6} \cmidrule(lr){7-8} \cmidrule(lr){9-10} \cmidrule(lr){11-12}
\textbf{Variant} & \textbf{Test} & \textbf{All} & \textbf{Non-} & \textbf{All} & \textbf{Non-} & \textbf{All} & \textbf{Non-} & \textbf{All} & \textbf{Non-} & \textbf{All} & \textbf{Non-} \\
 & \textbf{Set} & \textbf{Queries} & \textbf{Zero} & \textbf{Queries} & \textbf{Zero} & \textbf{Queries} & \textbf{Zero} & \textbf{Queries} & \textbf{Zero} & \textbf{Queries} & \textbf{Zero} \\
\midrule
Zero-Shot & A & 0.447 & 0.579 & 0.494 & 0.640 & 0.480 & 0.611 & 0.504 & 0.608 & 0.456 & 0.591 \\
baseline & B & 0.717 & 0.850 & 0.746 & 0.864 & 0.728 & 0.864 & 0.719 & 0.853 & 0.726 & 0.882 \\
\addlinespace[3pt]
\multirow{2}{*}{DA (Base)} & A & \textbf{0.675} & \underline{\textbf{0.752}} & \underline{\textbf{0.682}} & \textbf{0.750} & 0.599 & 0.705 & \textbf{0.651} & 0.702 & \textbf{0.669} & \textbf{0.726} \\
 & B & 0.812 & 0.920 & 0.842 & 0.954 & \textbf{0.816} & \textbf{0.946} & 0.816 & 0.925 & 0.828 & \textbf{0.939} \\
\addlinespace[3pt]
\multirow{2}{*}{DA + Back-Trans} & A & 0.651 & 0.730 & 0.645 & 0.734 & \textbf{0.618} & \textbf{0.718} & 0.637 & \textbf{0.710} & 0.650 & 0.710 \\
 & B & 0.823 & 0.932 & \underline{\textbf{0.870}} & \underline{\textbf{0.965}} & 0.771 & 0.915 & \textbf{0.827} & \textbf{0.937} & 0.811 & 0.919 \\
\addlinespace[3pt]
\multirow{2}{*}{DA + Paraph.} & A & 0.595 & 0.691 & 0.633 & 0.705 & 0.523 & 0.643 & 0.573 & 0.670 & 0.582 & 0.695 \\
 & B & 0.817 & \textbf{0.947} & 0.810 & 0.918 & 0.671 & 0.796 & 0.764 & 0.866 & 0.803 & 0.891 \\
\addlinespace[3pt]
\multirow{2}{*}{DA + Full Augm.} & A & 0.608 & 0.696 & 0.623 & 0.685 & 0.528 & 0.636 & 0.588 & 0.673 & 0.585 & 0.688 \\
 & B & \textbf{0.824} & 0.934 & 0.841 & 0.933 & 0.706 & 0.837 & 0.791 & 0.897 & \textbf{0.831} & 0.921 \\
\bottomrule
\end{tabular}%
}
\end{table}
 
\subsection{Test Set A: Control-to-Metric Association}
Domain adaptation improves performance consistently over the zero-shot baseline for all models on Test Set A, with the most notable observation being that training on the unaugmented dataset (DA Base) yields the best results for four out of five architectures. The largest absolute gain is achieved by the \textit{all-mpnet-base-v2 model}, which improves from 0.447 to 0.675 (an increase of 0.228 nDCG@10 points) when trained on the Base dataset. Similarly, \textit{multi-qa-mpnet-base-dot-v1} achieves the highest overall score on this test set with its Base variant, reaching 0.682 (an increase of 0.188 points compared to the baseline). 
The only exception to this trend is \textit{distilroberta-v1}, which achieves its highest score when the data is augmented with back-translation, yielding 0.618 (an improvement on its baseline score of 0.480 by 0.138). For the other models, data augmentation strategies generally resulted in a slight drop in performance on this specific task compared to the Base variant.
After fine-tuning, the gap between all-queries and non-zero performances is much smaller. To understand why this is the case, we must examine how the two averages work. Zero-shot models perform poorly on many queries, scoring exactly zero. Domain adaptation solves this issue. It helps the model identify relevant documents for queries that it had previously missed. This effectively transforms a zero score into a positive one.

\subsection{Test Set B: Cross-Standard Control Alignment}
Fine-tuning also improves all architectures on Test Set B, but the pattern differs from that on Test Set A in two key ways. Firstly,  the absolute gains are smaller, primarily because the zero-shot baselines for this task were already quite high (ranging from 0.717 to 0.746). Secondly, in contrast to Test Set A, data augmentation is highly beneficial here.
\textit{multi-qa-mpnet-base-dot-v1}, trained on the back-translation dataset, achieves the highest absolute score of all models and test sets, with remarkable results of 0.870 for all queries (+0.124 over its baseline) and 0.965 for non-zero queries (+0.101). Back-translation is also the optimal strategy for \textit{MiniLM-L12-v2}, achieving a score of 0.827. 
Conversely, \textit{all-mpnet-base-v2} and \textit{distilbert-cos-v1} achieve their best Test Set B scores overall using the fully augmented dataset.  \textit{distilroberta-v1} continues to exhibit distinct behavior, achieving its peak performance on Test Set B using the unaugmented Base dataset (0.816).
 
\subsection{The Effect of Augmentation Strategies}
The results show that the effectiveness of data augmentation depends on the nature of the evaluation task. For the control-to-metric association evaluated on Test Set A, the objective is to associate technical metrics with verbose legal mandates, and it appears that augmentation introduces noise. Simple semantic pairs from the base scenario provide the clearest signal and result in the best performance across nearly all architectures.

Cross-standard control association, evaluated on Test Set B, involves matching legal mandates formulated by different regulatory bodies with distinct lexicons, and augmentation becomes valuable.
Specifically, Back-Translation emerges as a highly robust strategy, propelling \textit{multi-qa-mpnet-base-dot-v1} and \textit{MiniLM-L12-v2} to their peak performance. The lexical variety introduced by intermediate translation probably helps the models generalize across the different phrasings typical of various security frameworks.  The Full Augmentation scenario also performs well here, showing that exposing the models to a broader vocabulary during training improves their performance on cross-standard association tasks.

\section{Conclusions}
This paper addresses the problem of automating compliance mapping between cloud security frameworks using fine-tuned Sentence Transformers. We construct a training corpus of 3,499 semantic pairs by aggregating seven heterogeneous standards around the Cisco CCF as a hub, and expand it via back-translation and LLM-based paraphrasing, yielding up to 13,996 training samples across four scenarios.

All fine-tuned models outperform their zero-shot baselines on both test sets. 

For the control-to-metric association task,  domain adaptation leads to improvements of up to 0.228 nDCG@10 over the baseline. These results extend our previous work~\cite{emerald1, emerald2} and support the role of domain-specific paired data as a key driver of performance gains.

For the control-to-control association task, the best-performing model (\textit{multi-qa-mpnet-base-dot-v1} trained with back-translation) achieves 0.870 nDCG@10 across all queries and 0.965 when considering only queries with at least one relevant match. This suggests that the model is able to align framework pairs not observed during training.

The results also highlight differentiated effects of data augmentation. While the unaugmented base dataset performs best for linking technical metrics to controls, augmentation strategies, particularly back-translation, are important for achieving stronger results in cross-standard mapping. The \textit{distilroberta-v1} model shows a distinct pattern, benefiting from back-translation on Test Set A but performing better with the unaugmented dataset on Test Set B.

All training data, fine-tuned models, and evaluation scripts are available in the EMERALD project repository.\footnote{\url{https://git.code.tecnalia.dev/emerald/public/components/mari/mari}}

In the future, we will incorporate systematic quality control into the augmentation pipeline. Rather than relying solely on manual spot-checks, we will use explainability methods, such as Integrated Gradients, to compare token-level attributions between the original and augmented pairs. This will enable us to identify and filter or regenerate low-quality augmentations that harm performance.


Another point is also to examine the scalability of the approach to denser evaluation settings  and extend the training corpus to additional compliance frameworks. Moreover, we plan to evaluate newer Sentence Transformer architectures and explore open-weight generative models, such as Llama 3 or Mistral, to transition to Retrieval-Augmented Generation (RAG) pipelines.

\section*{Acknowledgements}
This work is partially supported by EMERALD - Evidence Management for Continuous Certification as a Service in the Cloud (101120688) under the Horizon Europe program funded by the EU.

\bibliography{bibliography}
\bibliographystyle{elsarticle-harv}

\clearpage
\newpage
\end{document}